\documentclass[letterpaper, 10 pt, conference]{ieeeconf}  

\IEEEoverridecommandlockouts                              

\usepackage{graphics} 
\usepackage{epsfig} 
\usepackage{amsmath} 
\usepackage{cite}
\usepackage{caption}
\usepackage{multirow}
\usepackage{xcolor}
\usepackage{hyperref}
\usepackage{pifont}
\usepackage{array} 
\usepackage{latexsym}
\usepackage{amssymb}

\newcolumntype{P}[1]{>{\centering\arraybackslash}p{#1}}

\title{\LARGE \bf
Semi-Supervised Monocular Depth Estimation with Left-Right Consistency Using Deep Neural Network
}

\author{Ali Jahani Amiri, Shing Yan Loo, and Hong Zhang}

\begin{document}

\maketitle
\thispagestyle{empty}
\pagestyle{empty}

\begin{abstract}
There has been tremendous research progress in estimating the depth of a scene from a monocular camera image.  Existing methods for single-image depth prediction are exclusively based on deep neural networks, and their training can be unsupervised using stereo image pairs, supervised using LiDAR point clouds, or semi-supervised using both stereo and LiDAR. In general, semi-supervised training is preferred as it does not suffer from the weaknesses of either supervised training, resulting from the difference in the camera’s and the LiDAR’s field of view, or unsupervised training, resulting from the poor depth accuracy that can be recovered from a stereo pair.  In this paper, we present our research in single-image depth prediction using semi-supervised training that outperforms the state-of-the-art.  We achieve this through a loss function that explicitly exploits left-right consistency in a stereo reconstruction, which has not been adopted in previous semi-supervised training.  In addition, we describe the correct use of ground truth depth derived from LiDAR that can significantly reduce prediction error.  The performance of our depth prediction model is evaluated on popular datasets, and the importance of each aspect of our semi-supervised training approach is demonstrated through experimental results. Our deep neural network model has been made publicly available.\footnote{Source code is available at {\color{magenta} \hyperref[https://github.com/a-jahani/semiDepth]{https://github.com/a-jahani/semiDepth}}}.

\end{abstract}

\section{INTRODUCTION}

Single-image depth estimation is an important yet challenging task in the field of robotics and computer vision. A solution to this task can be used in a broad range of applications such as localization of the robot poses\cite{loo2018cnn, Yang_2018_ECCV}, 3D reconstruction in simultaneous localization and mapping\cite{tateno2017cnn}, collision avoidance\cite{chakravarty2017cnn}, and grasping\cite{rao2010grasping}. With the rise of deep learning, notable achievements in terms of accuracy and robustness have been obtained in the study of single image depth estimation, and methods of supervised,  unsupervised,  and semi-supervised have been proposed. 

Supervised methods in single-image depth estimation use ground truth derived from LiDAR data. It is time-consuming and expensive to obtain dense ground-truth depth, especially for the outdoor scenes. LiDAR data is also sparse relative to the camera view, and it does not share the same field of view with the camera in general. Consequently, supervised methods are unable to produce meaningful depth estimation in the non-overlapping regions with the image.
In contrast, unsupervised methods learn dense depth prediction using the principle of reconstruction from stereo views; hence depth can be estimated for the entire image. However, the accuracy of unsupervised depth estimation is limited by that of stereo reconstruction.\\


 \begin{figure} 
        \centering
        \includegraphics{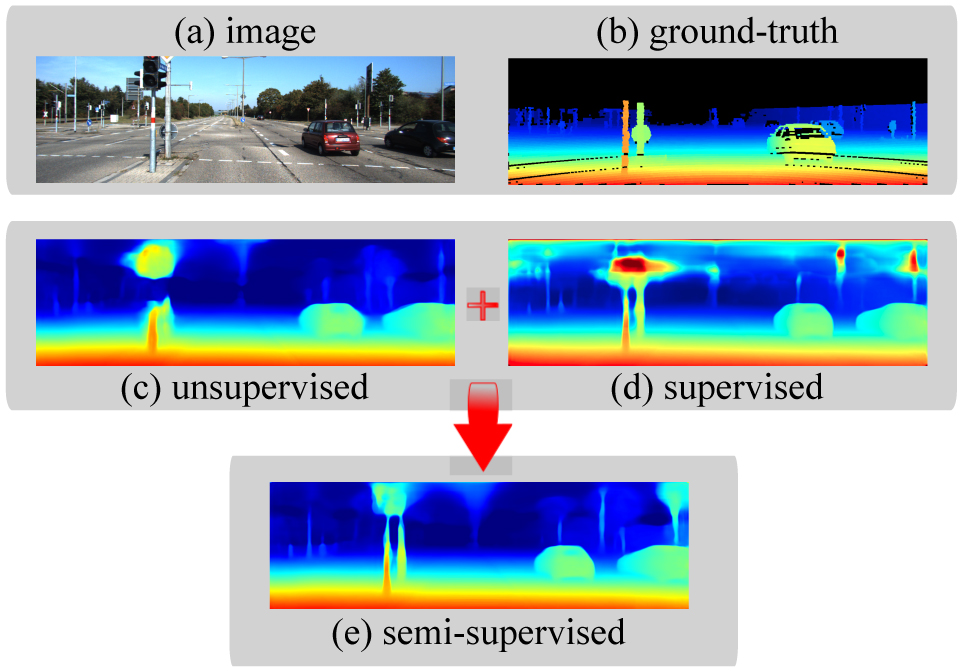}
        \caption{Using stereo only (c) leads to the noisy depth map. Using LiDAR only (d) results in inaccurate for the top part of the image because there is no ground-truth available. Our semi-supervised method (e) fuses both LiDAR and Stereo and can predict depth more accurately. Ground truth LiDAR (b) has been interpolated for visualization purpose.}
        \label{fig:loss}
    \end{figure}
In this paper, we present our research in single-image depth prediction using semi-supervised training that outperforms the state-of-the-art. We propose a novel semi-supervised loss that uses left-right consistency term originally proposed in \cite{godard2017unsupervised}. Our network uses LiDAR data for supervised training and rectified stereo images for unsupervised training, and in the testing phase, our network takes only one image to perform depth estimation. 

Another focus of our study is the impact of ground truth depth information on the training of our model, when network training is performed with the projected raw LiDAR data and the annotated depth map recently provided by KITTI \cite{Uhrig2017THREEDV}, respectively. We discover that the commonly used projected raw LiDAR contains noisy artifacts due to the displacement between the LiDAR and the camera, leading to poor network performance. In contrast, we use the more reliable preprocessed annotated depth map for training, and we are able to achieve a significant reduction of prediction error.

In summary, we propose in this paper a semi-supervised deep neural network for depth estimation from a single image, with state-of-the-art performance.  Our work makes the following three main contributions.
\begin{itemize}
\item  We show the importance of including a left-right consistency term in the loss function for performance optimization in semi-supervised single-image prediction.
\item We provide empirical evidence that training with the annotated ground truth derived from LiDAR leads to better depth prediction accuracy than with the raw LiDAR data as ground truth.
\item  We make our semi-supervised deep neural network - based on the popular Monodepth \cite{godard2017unsupervised} architecture - available to the community.
\end{itemize}

The rest of this paper is organized as follows.  In Section II, we will review related works to our research, and in Section III we will present our proposed neural network model for single-image depth estimation.   Experimental evaluation of our proposed model is provided in Section IV, and conclusion of our work in Section V.

\section{Related works}
Over the past few years, numerous deep learning based methods have been proposed for the problem of single-image depth estimation. We can roughly divide these deep methods into three categories:  supervised, unsupervised, and semi-supervised.

\subsection{Supervised}\label{sec:supervised}

Supervised methods use ground truth depth, usually from LiDAR in outdoor scenes, for training a network. Eigen et.al.\cite{eigen2014depth} was one of the first who used such a method to train a convolutional neural network. First, they generate the coarse prediction and then use another network to refine the coarse output to produce a more accurate depth map. Following \cite{eigen2014depth}, several techniques have been proposed to improve the accuracy of convolutional neural networks such as CRFs \cite{li2015depth}, inverse Huber loss as a more robust loss function \cite{laina2016deeper}, joint optimization of surface normal and depth in the loss function \cite{wang2015designing,hu2018revisiting,qi2018geonet}, fusion of multiple depths maps using Fourier transform\cite{lee2018single}, and formulation of depth estimation as a problem of classification\cite {fu2018deep}.

\subsection{Unsupervised}\label{sec:unsupervised}
 To avoid laborious ground truth depth construction, unsupervised methods based on stereo image pairs have been proposed\cite{xie2016deep3d}. Garg et al. \cite{garg2016unsupervised} demonstrated an unsupervised method in which the network is trained to minimize the stereo reconstruction loss; i.e., the loss is defined such that the reconstructed right image (i.e., obtained by warping the left image using the predicted disparity) matches the right image. Later on, Godard et al.\cite{godard2017unsupervised} extended the idea by enforcing a left-right consistency that makes the left-view disparity map consistent with the right-view disparity map. The unsupervised training of our model is based on \cite{godard2017unsupervised}. Given a left view as input, the model in\cite{godard2017unsupervised} outputs two disparities of the left view and the right view, while we are outputting only one depth map for one input image in the form of inverse depth instead of disparity. As a result, we treat both left and right images equivalently which allows us to eliminate the overhead of the post-processing step in \cite{godard2017unsupervised}. By making these changes, our unsupervised model outperforms \cite{godard2017unsupervised} as will be discussed in Section \ref{sec:Experiments}.

 \subsection{Semi-Supervised}\label{sec:monoslam}

Unlike unsupervised methods, there has not been much work on semi-supervised learning of depth. Luo et al. \cite{luo2018single} and Guo et al.\cite{Guo_2018_ECCV} proposed a method that consists of multiple sequential unsupervised and supervised training stages; hence their method could be categorized as a semi-supervised method although they did not use LiDAR and stereo images at the same time in training.
Closest to our work is Kuznietsov et al. \cite{kuznietsov2017semi} who proposed adding the supervised and unsupervised loss term in the final loss together resulting in using LiDAR and stereo at the same time in training. One of the main differences between\cite{kuznietsov2017semi} and ours is that we have the left-right consistency term first proposed by \cite{godard2017unsupervised}. Having this term makes the prediction consistent between left and right. Another difference is that their supervised loss term was directly defined on the depth values whereas we defined it on inverse depth instead. As discussed in \cite{kuznietsov2017semi}, a loss term on depth values makes the training unstable because of the high gradients in the early stages of the training. To remedy the situation, Kuznietsov et al. proposed to gradually fade in the supervised loss to achieve convergence whereas our method does not have this problem and does not need to fade in supervised or unsupervised loss terms. In Section \ref{sec:result}, we show qualitatively and quantitatively that we can obtain better accuracy than \cite{kuznietsov2017semi}, which is considered the state-of-the-art in semi-supervised single image depth estimation, as the result of the above considerations. 
 
 \begin{figure*} [!h]
        \centering
        \includegraphics[scale=0.9]{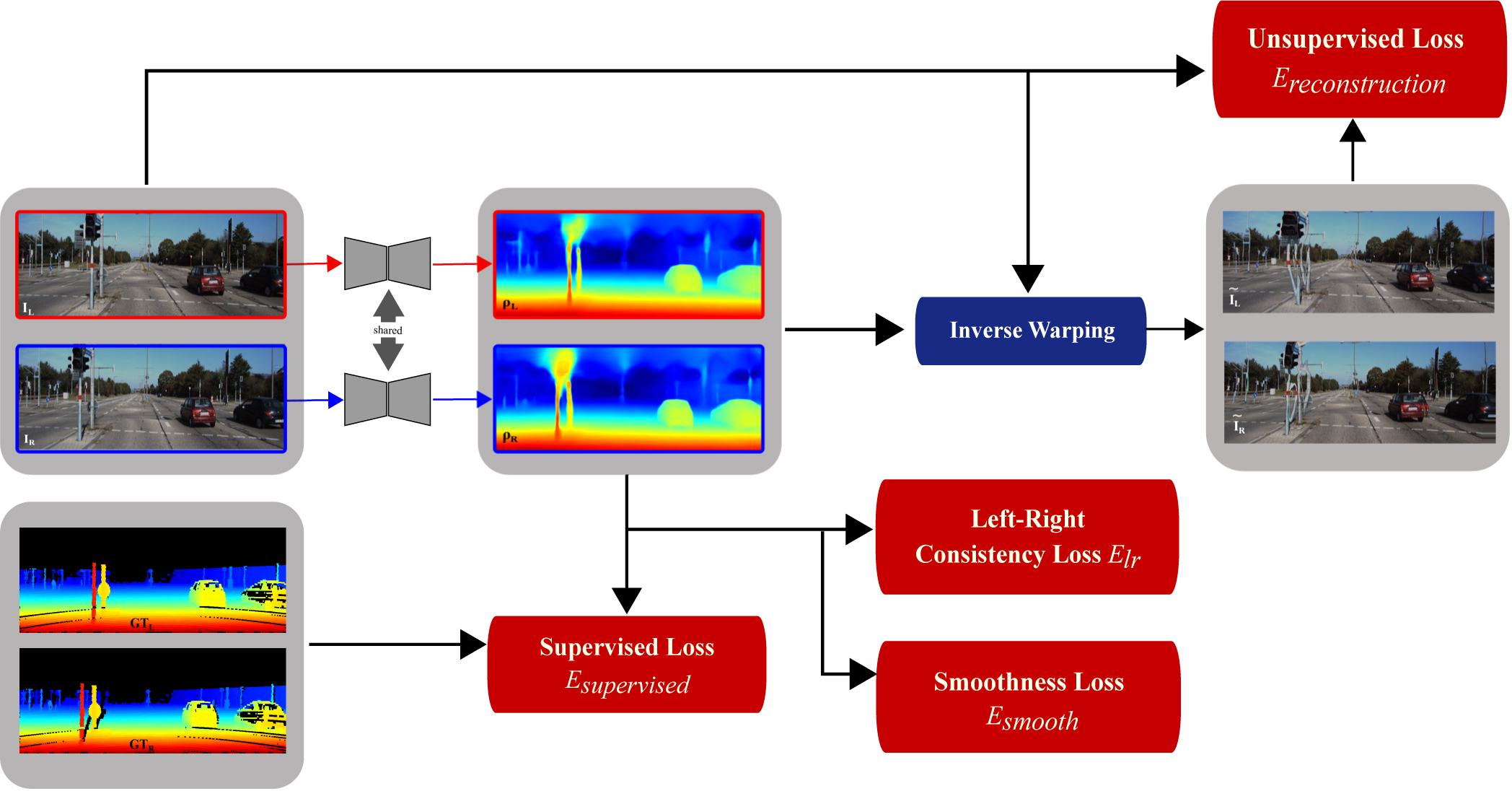}
        \caption{Overview of the schematic of our proposed loss. There are 4 terms in our loss $E_{reconstruction}$, $E_{supervised}$, $E_{lr}$, $E_{smooth}$, $E_{supervised}$. Subscript $L$ and $R$ refers to left and right image, respectively. $\rho$ refers to output of our network inverse depth. We use bilinear sampler in the inverse warping function.}
        \label{fig:loss}
    \end{figure*}

\section{Method}
 Our approach is based on Monodepth proposed by Godard et al.\cite{godard2017unsupervised}. Their work is unsupervised and only uses rectified stereo images in training. In this paper, we extend their work and add ground-truth depth data as additional supervision training data. To the best of our knowledge, we are the first one to use left-right consistency proposed by Godard \cite{godard2017unsupervised} in a semi-supervised framework of single image depth estimation. Fig. \ref{fig:loss} shows the different loss terms we use in our training phase, to be described in detail in the next section. 

\subsection{Loss Terms}
Similar to \cite{godard2017unsupervised}, we define $L_{s}$ for each output scale s. Hence the total loss is defined as $L_{total}=\sum_{s=1}^{4}{L_{s}}$. 

\begin {equation}
L_{s} = \lambda_{1}E_{reconstruction} +\lambda_{2}E_{lr}  +\lambda_{3}E_{supervised} +\lambda_{4}E_{smooth}  
\end{equation}
where $\lambda_{i}$ are scalars and the $E$ terms are defined below:

\subsubsection{Unsupervised Loss $E_{reconstruction}$}
We use photometric reconstruction loss between left and right image. Similar to other unsupervised methods, we assume photometric constancy between left-right images. Inverse warping has been used to get the estimated left/right image and then the estimated image is compared with its corresponding real image. In the inverse warping, bilinear sampler is used to make the pipeline differentiable. For comparison, we use the combination of the structural similarity index (SSIM) and L1 used by Godard et al\cite{godard2017unsupervised}, and the ternary census transform used in ~\cite{meister2018unflow,zabih1994non,stein2004efficient}. SSIM and the ternary census transform can compensate for the gamma and illumination change to some extent and result in improved satisfaction of the constancy assumption. Our unsupervised photometric image reconstruction loss term $E_{u}$ is defined as follows:
\begin{equation}
\begin{split}
E_{reconstruction}= &\sum_{k \in \{l,r\}}f(I^{k},\tilde{I}^{k})\\
f(I,\tilde{I})=&\frac{1}{N}\sum\limits_{i,j} \alpha_{1}*\frac{1- SSIM(I_{ij},\tilde{I}_{ij})}{2} + \\
&\alpha_{2} * ||I_{ij} - \tilde{I}_{ij} ||_{1}+ \\
 &\alpha_{3} * census(I_{ij},\tilde{I}_{ij})
\end{split}
\end{equation}
where $I^{l}$, $I^{r}$, $\tilde{I}^{l}$, and $\tilde{I}^{r}$  are the left image, right image and their reconstructed images, respectively. N is the total number of pixels. $\alpha_{1}$, $\alpha_{2}$, and $\alpha_{3}$ are scalars that define the contribution of each term to the total reconstruction loss.\\
\\

\subsubsection{Left-Right Consistency Loss $E_{lr}$}To ensure the equal contribution of both left and right images in the network training, we feed left and right images independently to the network, and then we jointly optimize the output of the network such that the predicted left and right depth maps are consistent. As explained in \cite{godard2017unsupervised}, left-right consistency loss attempts to make the inverse depth of the left (or right) view the same as the projected inverse depth of the right (or left) view. This type of loss is similar to forward-backward consistency for optical flow estimation \cite{meister2018unflow}. We define our left-right consistency loss as follows:\\
\begin{equation}
E_{lr}=\frac{1}{N}\sum\limits_{i,j} || \rho_{ij}^{l} - \rho_{ij+d_{ij}^{l}}^{r} ||_{1} + || \rho_{ij}^{r} - \rho_{ij+d_{ij}^{r}}^{l} ||_{1},
\end{equation}
where $\rho_{l}$ and $\rho_{r}$ are the predicted inverse depth for left and right images, respectively. $d^{l}$ and $d^{r}$ are predicted disparities correspond to left and right images, respectively. The conversion of inverse depth $\rho$ to disparity $d$ is calculated using \eqref{eq:bfd}:

\begin{equation}\label{eq:bfd}
    d = baseline * f * \rho ,
\end{equation}
where $f$ is the focal length of the camera.
\\
\begin{figure*} [t]
        \centering
        \includegraphics[scale=0.9]{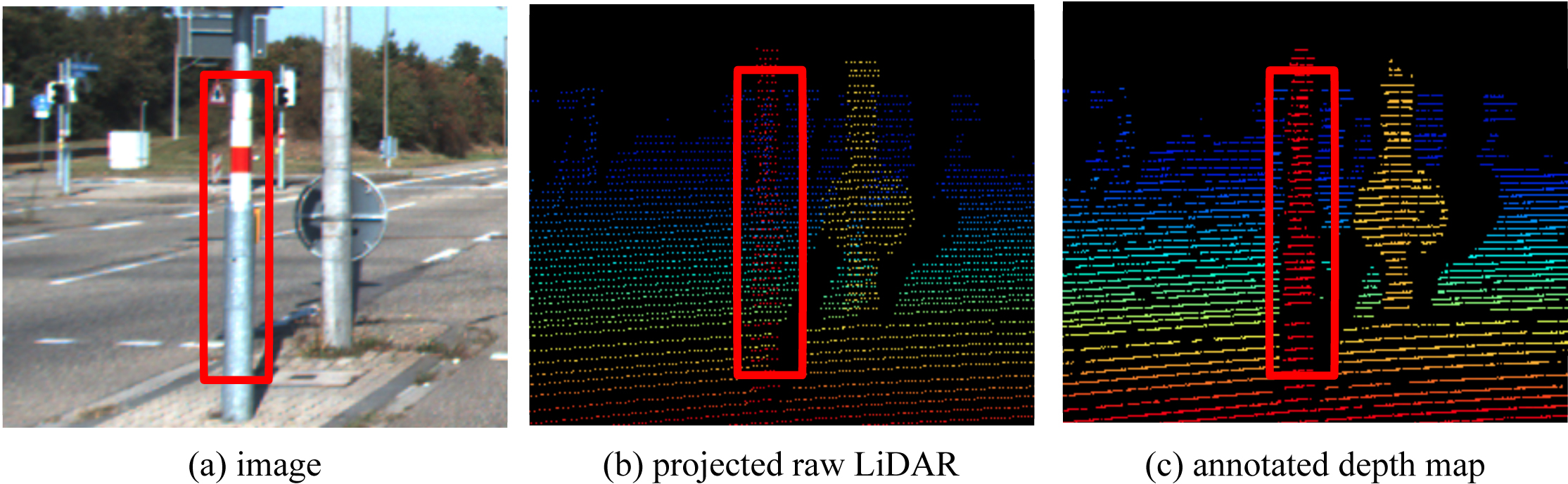}
        \caption{Qualitative comparison between (b) projected raw LiDAR containing occlusion artifacts due to the displacement between the camera and LiDAR and (c) annotated depth map without any occlusion artifact. We use annotated depth map dataset (c) for our training and evaluation. (b) shows the erroneous depth values for points (green pixels among red for the pole bounded by the red rectangle) that are occluded from the camera point of view but not LiDAR point of view.}
        \label{fig:occlusion}
\end{figure*}

\subsubsection{Supervised Loss $E_{s}$}
The supervised loss term measures the difference between the ground truth inverse depth $Z^{-1}$ and the predicted inverse depth $\rho$ for the points $\Omega$ where the ground truth is available. \\
\begin{equation}
E_{supervised}= \sum\limits_{k \in \{l,r\}}\frac{1}{M_{k}}\sum\limits_{i,j \in \Omega_{k}} {||\rho_{ij}^{k}- {Z^{-1}}_{ij}^{k}  ||_{1}} \\
\end{equation}
where $\Omega_{l}$ and $\Omega_{r}$ are the points where the ground truth depths are available for the left and right images, respectively.  $M_{l}$ and $M_{r}$ are the total number of the pixels that ground truth is available for left and right images, respectively.\\

\subsubsection{Smoothness Loss $E_{smooth}$}
As suggested in \cite{godard2017unsupervised, kuznietsov2017semi}, the smoothness loss term is a regularization term that encourages the inverse depth to be locally smooth with a $L_{1}$ penalty on inverse depth gradients. We define our smoothness regularization term as:

\begin{equation}
    E_{smooth}= \frac{1}{N}\sum_{k\in \{l,r\}} \sum\limits_{i,j}|\partial_{x}\rho_{ij}^k|e^{-|\partial_{x}I_{ij}^{k}|}+ |\partial_{y}\rho_{ij}^k|e^{-|\partial{y}I_{ij}^{k}|}  
\end{equation}
 Since the depth is not continuous around object boundaries, this term encourages the neighbouring depth values to be similar in low gradient image regions and dissimilar otherwise.
 
\section{Experiments} \label{sec:Experiments}
\begin{table*}
\centering
\begin{tabular}{|c||c||c||c|c|c|c||c|c|c|}
\hline
\multirow{2}{*}{method}                & \multirow{2}{*}{type}& \multirow{2}{*}{Dataset}&  \multicolumn{4}{c||}{lower is better}&\multicolumn{3}{c|}{higher is better}    \\ \cline{4-10} &                       &             &  Abs Rel    & Sq Rel      & RMSE        & RMSE$_{log}$  &$\delta < 1.25$ & $\delta<1.25^{2}$ & $\delta <1.25^{3}$    \\ \hline
raw LiDAR            &        -            &  -   &    0.010    &     0.126  &     1.209         &     0.054    &    0.993     &    0.996    &   0.998   \\ \hline
DORN\cite{fu2018deep}             &        S            &  \textit{K}   &   \textbf{0.080}    &     \textbf{0.332}  &     \textbf{2.888}   &     \textbf{0.120}   &    \textbf{0.938}     &    \textbf{0.986}    &   \textbf{0.995}   \\
\textbf{SemiDepth(Ours)}                                  &        S            &  \textit{K}   &     0.096        &    0.552         &       3.995      &         0.152    &       0.892      &      0.972       &     0.992     \\  \hline

Monodepth \cite{godard2017unsupervised}(Resnet50)&  U    &  \textit{C+K}  &   0.085     &    0.584   &  3.938      &     0.135   &       0.916 &   \textbf{0.980}   &   0.994  \\
MonoGAN \cite{aleotti2018generative}(Resnet50)&  U    &  \textit{C+K}  &   0.096     &    0.699   &  4.236      &     0.150   &       0.899 &    0.974   &   0.992  \\
\textbf{SemiDepth(Ours)}                                   &     U    &  \textit{C+K}  &    \textbf{0.082}   &     \textbf{0.551}  &  \textbf{3.837}      &     \textbf{0.134} & \textbf{0.920}    &     \textbf{0.980}   &  \textbf{0.993}   \\\hline

Kuznietsov et. al. \cite{kuznietsov2017semi}   &     Semi    &  \textit{I+K}  &   0.089     &   0.478     &  3.610        &   0.138    &    0.906    &    0.980    &   0.995  \\ 
SVSM FT \cite{luo2018single}           &        Semi         &  \textit{I+F+K}    &   \textbf{0.077}    &   \textbf{0.392}   &  3.569      &     0.127    &    0.919     &      0.983  &    0.995 \\
\textbf{SemiDepth (full) (Ours) }       &        Semi         &  \textit{C+K}    &    0.078  &   0.417  &     \textbf{3.464}    &    \textbf{0.126}       &    \textbf{0.923}   &    \textbf{0.984}        &    \textbf{0.995}      \\  \hline
\end{tabular}
\caption{Quantative evaluation on 652 (93.5\%) test images of Eigen Split from the KITTI Dataset. We use official annotated depth map dataset as ground truth instead of noisy projected raw LiDAR. U, S, Semi means unsupervised, supervised and semi-supervised training, respectively. Results of \cite{godard2017unsupervised, aleotti2018generative} are achieved without the post-processing step. \textit{I}, \textit{C}, \textit{K}, and \textit{F} refer to ImageNet\cite{imagenet_cvpr09}, Cityscapes\cite{Cordts2016Cityscapes}, KITTI\cite{Uhrig2017THREEDV} and FlyingThings3D datasets, respectively. \textit{I} indicates that an encoder is initialized with a pre-trained model trained on ImageNet. All evaluations are using crop from \cite{garg2016unsupervised}. Depth is capped at 80.0 meters.}
\label{tab:result}
\end{table*}

\begin{table*}[!h]
\centering
\begin{tabular}{| P{1.5cm}| P{1.5cm}||P{1.5cm}|||c|c|c|c|}
\cline{1-3}
\multicolumn{2}{|c||}{Training Data}&  Loss Term &\multicolumn{4}{|c}{} \\ \hline 
 \cline{1-3}
    Projected Raw LiDAR & Annotated Depth map   & Left-Right Consistency  &  \multirow{2}{*}{Abs Rel}   & \multirow{2}{*}{Sq Rel} & \multirow{2}{*}{RMSE} & \multirow{2}{*}{RMSE$_{log}$}\\
    \hline
      \ding{51}         &                        &    \ding{51}               &    0.120  &  1.154   &  5.614 & 0.204\\     
     
               &    \ding{51}                       &                &     0.110   & 0.973 &  5.373 & 0.191\\ 
              &    \ding{51}                       &    \ding{51}              &     \textbf{0.108}   & \textbf{0.949} &  \textbf{5.369} & \textbf{0.190} \\ \hline

\end{tabular}
\caption{The effect of the left-right cons:istency term and using annotated depth map in our semi-supuervised training. The results are evaluated on 200 images of KITTI Stereo 2015 split\cite{Menze2018JPRS}. The second and the third row show that exploiting left-right consistency helps achieving better accuracy. The first and the third row show training on annotated depth map significantly reduces error. The result from our method is shown in bold.} 
\label{tab:changes}
\end{table*}


      

 For comparison, we use the popular Eigen split\cite{eigen2014depth} in KITTI dataset\cite{Uhrig2017THREEDV} that has been used in the previous methods. Using this split, we notice the same problem mentioned by Aleotti et al.\cite{aleotti2018generative} that, when LiDAR points are projected into the camera space, an artifact results around objects that are occluded in the image but not from the LiDAR point of view. This is due to the displacement between the LiDAR and the camera sensors. Recently Uhrig et al. \cite{Uhrig2017THREEDV} provided preprocessed annotated depth maps of KITTI by a preprocessing step on projected raw LiDAR data. They used multiple sequences, left-right consistency checks, and untwisting methods to carefully filter out outliers and densify projected raw LiDAR point clouds. Fig. \ref{fig:occlusion} shows the occlusion artifact in raw projected LiDAR and the corresponding annotated depth map dataset provided by \cite{Uhrig2017THREEDV}. Since the occlusion artifact is filtered out in the annotated depth ground truth, we train our model with this more accurate ground truth. The first and the third row of Table \ref{tab:changes} show the effect of the training network with the projected raw LiDAR versus the annotated ground truth.
 
In the rest of the experiments, we evaluate our method based on the official KITTI annotated depth map rather than noisy projected raw LiDAR. Table \ref{tab:result} contains the quantitative evaluation of the projected raw LiDAR based on the provided annotated depth map ground truth if a depth value of a pixel exists in the both annotated depth map and projected raw LiDAR (54.89\% of the LiDAR points have been evaluated). The large error for projected raw LiDAR suggests that raw LiDAR is not as accurate as annotated depth maps.

\subsection{Evaulation Metrics}
We use the standard metrics used by previous researchers.\cite{eigen2014depth,kuznietsov2017semi,godard2017unsupervised}. Specifically, we use RMSE, $\text{RMSE}_{log}$, absolute relative difference (Abs Rel), squared relative difference (Sq Rel), and the percentage of depths ($\delta$) within a certain threshold distance to its ground truth.   



\subsection{Implementation Details}
We train our network from scratch using Tensorflow \cite{abadi2016tensorflow}. Our network and training procedure are identical to the Resnet50 network used by Godard et al.\cite{godard2017unsupervised} except for the decoder part in which we have one output instead of two for each scale. As \cite{godard2017unsupervised} all inputs are resized to 256*512. The output of the network,i.e., inverse depth, is limited to 0 to 1.0 using the sigmoid function. We use Adam optimiser \cite{kingma2014adam} with $\beta _{1} = 0.9$, $\beta _{2} = 0.999$, and $\epsilon = 10^{-8}$ with initial learning rate of $\lambda = 10^{-4}$, and that remains constant for the first 15 epochs and being halved every 5 epochs for the next 10 epochs for a total of 25 epochs. The hyperparameters for loss are chosen as $\lambda _{1}=1$, $\lambda _{2}= 1.0$, $\lambda _{3} = 150.0$, $\lambda _{4} = 0.1$, $\alpha _{1} = 0.85$, $\alpha _{2} = 0.15$, and  $\alpha _{3}=0.08$.

\subsection{Results} \label{sec:result}

\begin{figure*} [!h]
        \centering
        \includegraphics[scale=0.85]{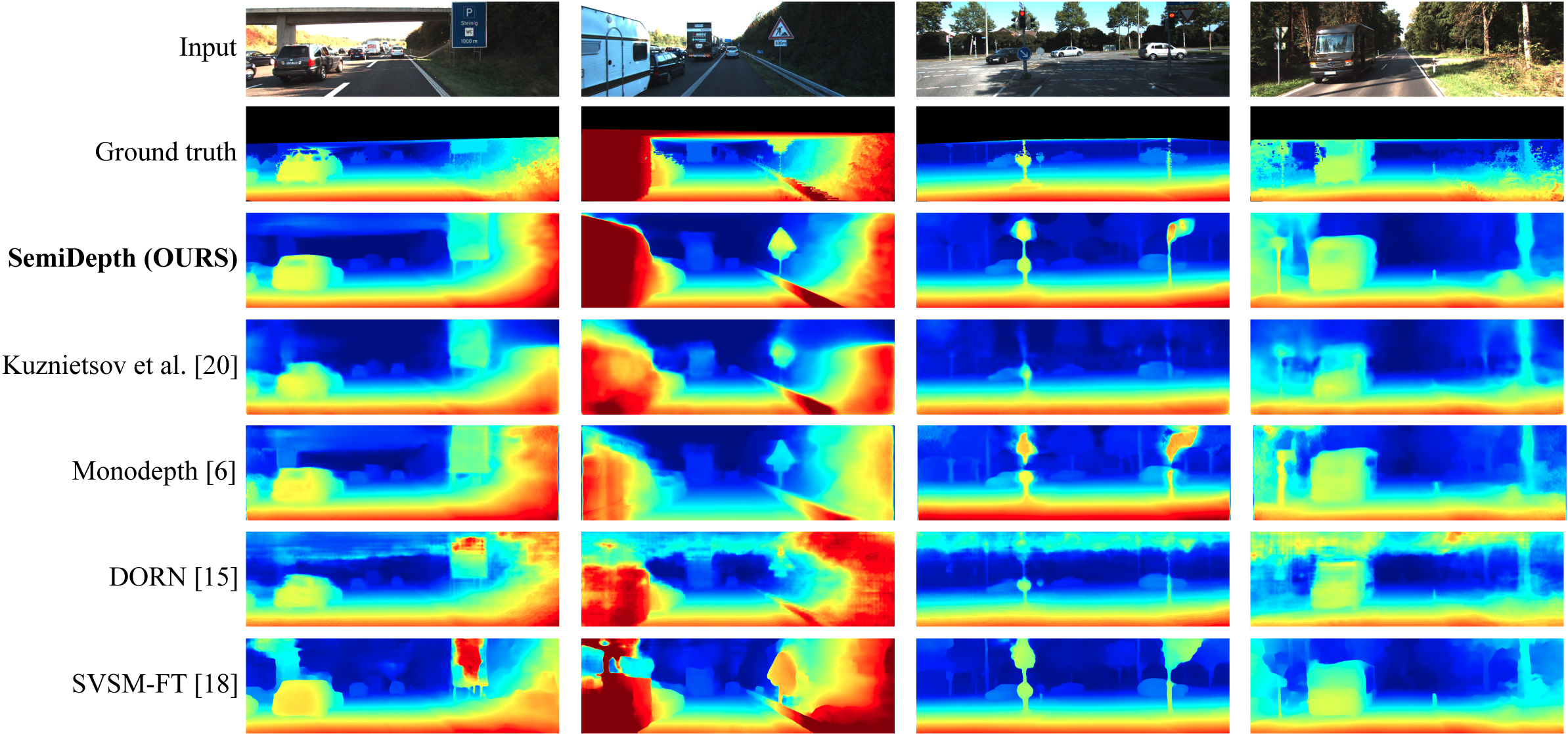}
        \caption{Qualitative comparison between state-of-the-art methods. We use interpolation in ground truth for visualization purpose.}
        \label{fig:result2}
    \end{figure*}

Table \ref{tab:result} shows the quantitative comparison with the state of the art methods in Eigen split using reliable annotated depth maps for training and testing. Although supervised methods, e.g., DORN\cite{fu2018deep} can achieve better quantitative performance according to some metrics than semi-supervised methods, they produce an inaccurate prediction of the top portion of the image, which can be seen in Fig. \ref{fig:result2}, where the LiDAR's field of view is different from that of the camera.

By treating left and right images equivalently and defining our loss symmetrically, we eliminate the post-processing step needed in \cite{godard2017unsupervised}. As shown in Table \ref{tab:result}, our unsupervised model outperforms our baseline unsupervised model \cite{godard2017unsupervised}. In addition, from Table \ref{tab:result} among the evaluated semi-supervised methods, our method outperforms \cite{kuznietsov2017semi}, considered the state-of-the-art, with respect to the majority of the performance metrics. To investigate in detail the effect of using left-right consistency term in the loss function and that of using the annotated LiDAR ground truth, the advantage of our method is confirmed in Table \ref{tab:changes}, where 200 images of KITTI Stereo 2015 split\cite{Menze2018JPRS} were used in this controlled experiment.
 
\section{conclusion}
In this paper, we have presented our approach to semi-supervised training of a deep neural network for single-image depth prediction.  Our network uses a novel loss function that uses the left-right consistency term, which has not been used in the semi-supervised training of depth-prediction networks. In addition, we have explained and experimentally confirmed that, for optimal prediction result, in either supervised or semi-supervised training, careful use of the LiDAR data as the ground truth is important. Extensive experiments have been conducted to evaluate our proposed training approach, and we are able to achieve state-of-the-art performance in depth prediction accuracy.  Our network model, which is based on Monodepth that is popularly used within the robotics community, is available online for download.

\section*{Acknowledgments}
This work has been supported by NSERC Canadian Robotics Network (NCRN). We would like to thank Godard et al.\cite{godard2017unsupervised} for making their code publicly available.






\bibliographystyle{unsrt}

\bibliography{refs}

\addtolength{\textheight}{-12cm} 


\end{document}